\documentclass{article}

\usepackage{PRIMEarxiv}

\usepackage[ruled,vlined]{algorithm2e}
\usepackage[utf8]{inputenc} 
\usepackage[T1]{fontenc}    
\usepackage{hyperref}       
\usepackage{url}            
\usepackage{booktabs}       
\usepackage{amsfonts}       
\usepackage{nicefrac}       
\usepackage{microtype}      
\usepackage{lipsum}
\usepackage{fancyhdr}       
\usepackage{graphicx}       
\graphicspath{{media/}}     

\title{Model-Agnostic Meta-Learning for Natural
Language Understanding Tasks in Finance
}

\author{
    Bixing Yan \footnotemark[1] \space \footnotemark[2]\\
  Center for Data Science\\
  New Your University\\
  New York, NY 10012\\
  \texttt{by783@nyu.edu} \\
  \And
  Shaoling Chen \footnotemark[1] \\
  Center for Data Science\\
  New Your University\\
  New York, NY 10012\\
  \texttt{sc6995@nyu.edu} \\
  \AND
  Yuxuan He \footnotemark[1] \\
  Center for Data Science\\
  New Your University\\
  New York, NY 10012\\
  \texttt{yh2857@nyu.edu} \\
  \And
  Zhihan Li \footnotemark[1] \\
  Center for Data Science\\
  New Your University\\
  New York, NY 10012\\
  \texttt{zl2516@nyu.edu} \\
}

\begin{document}
\maketitle
\footnotetext[1]{The authors contributed equally and are presented in alphabetical order.}
\footnotetext[2]{Bixing Yan is the corresponding author.}

\begin{abstract}
Natural language understanding(NLU) is challenging for finance due to the lack of annotated data and the specialized language in that domain. As a result, researchers have proposed to use pre-trained language model and multi-task learning to learn robust representations. However, aggressive fine-tuning often causes over-fitting and multi-task learning may favor tasks with significantly larger amounts data, etc. To address these problems, in this paper, we investigate model-agnostic meta-learning algorithm(MAML) in low-resource financial NLU tasks. Our contribution includes: 1. we explore the performance of MAML method with multiple types of tasks: GLUE datasets, SNLI, Sci-Tail and Financial PhraseBank; 2. we study the performance of MAML method with multiple single-type tasks: a real scenario stock price prediction problem with twitter text data. Our models achieve the state-of-the-art performance according to the experimental results, which demonstrate that our method can adapt fast and well to low-resource situations.
\end{abstract}


\section{Introduction}

It has been a trading practice tradition to utilize textual data to improve modeling of the financial market dynamics\cite{xing_natural_2018}. Nowadays financial operators have access to a growing volume of information, provided by financial reports, news articles, press releases, etc. The enrichment of text sources has also lead to diverse types of unstructured and structured data, for example, social media websites like Twitter, Facebook, etc. are generating rich text content, which can be used as a supplement to support prediction. As a result, there have been increasing attempts to try to utilize deep learning methods on solving financial tasks, including financial opinion mining and question answering \cite{noauthor_fiqa_nodate}, financial sentiment analysis\cite{araci_finbert:_2019}, financial named entity recognition\cite{7009718} and other natural language understanding(NLU) tasks.

However, traditional deep neural network based methods faces several drawbacks. First, they often require vast amount of annotated data which requires high manual labeling cost. Second,  language model that trained on Wikitext or other general dataset may be ineffective in solving financial tasks \cite{araci_finbert:_2019} because text data in financial field often exhibits a different pattern compare to text data collected in other domain. Thus, aiming at solving this issue, researchers and investors in financial NLU field has shifted their attention to use transfer learning technique, i.e. to learn a general representation of financial text and adapt it to other new tasks.

Researchers have presented several approaches for transfer learning in Finance NLU field. One of the approach is FinBERT\cite{araci_finbert:_2019}, which exploits the powerful pre-trained language model, BERT\cite{devlin_bert:_2019} fine-tunes it using texts in financial field then uses it for new tasks. Further, another approach is to apply multi-task learning to representation learning, where \cite{liu_multi-task_2019} proved that BERT model could be improved with multi-task learning strategy as the MT-DNN model. It has achieved descent results on GLUE datasets. However, \cite{dou_investigating_2019} pointed out that multi-task learning may prefer tasks with significantly larger datasets than others and further suggested meta-learning algorithms for multiple types of low resource language understanding tasks. Meta-learning algorithms try to learn a meta-policy for updating model parameters or a good initialization that can be useful for fine-tuning on various tasks with minimal training data, which makes them promising alternatives to multi-task learning. Meta-learning has been proved useful in few-shot learning\cite{finn_model-agnostic_2017},single-type multi-tasks learning. Indeed, \cite{gu_meta-learning_2018} extends meta-learning algorithm for low-resource neural machine translation, framing low-resource translation as a meta-learning problem and adapting to low-resource languages based on multilingual high-resource language tasks.  

Inspired by these work, in this paper, we investigate the applications of meta-learning algorithms, specifically the Model-Agnostic Meta-Learning(MAML) algorithm\cite{finn_model-agnostic_2017}, to try to solve the fundamental representation learning issue in financial text data.

The main contribution of this paper is two-fold:
\begin{itemize}
\item We study the performance of MAML method with multiple types of tasks. We combine the MAML algorithm with MT-DNN model, train the model using four high-resource datasets, evaluate it on other low-resource datasets, and then adapt the model to Financial PhraseBank, a financial sentiment analysis dataset, where we achieve the state-of-the-art. Our experiments also justify the superior property in fast adaptation and over-fitting avoidance of the MAML model.

\item We study the performance of MAML method with multiple tasks in single type. We develop a few-shot learning method for the task of stock price movement prediction with news texts, and propose a competitive MAML-BERT model for stock price prediction.
\end{itemize}

The rest of the paper is structured as follows: Section 2 briefly describes the relevant literature in multi-task learning, meta-learning and financial natural language understanding tasks. Then, Section 3 introduces the methods we use: BERT model and MAML algorithm. In Section 4, we present multiple experiments being conducted, including datasets, implementations and their results. We conclude with Section 5 and discuss the future work in Section 6.

\section{Related Work}
\label{gen_inst}
In this section, we introduce the relevant literature in multi-task learning, meta-learning and two financial natural language understanding tasks including financial sentiment analysis and stock price prediction.
\subsection{Multi-Task Learning}

Multi-task learning (MTL) is a sub-field of machine learning, which exploits commonalities and differences across tasks and solves multiple learning tasks at the same time. Biologically, we often apply the knowledge we have acquired in related tasks to learning new tasks. For example, a baby first learns to recognize faces and can then recognize other objects by applying this knowledge. Similarly, multi-task learning can result in improved learning efficiency and prediction accuracy for the task-specific models, compared to training the models separately \cite{baxter_model_2000}.

Multi-task learning penalizes all complexity uniformly, and as a result, regularization induced by requiring an algorithm to perform well on a related task can be superior to regularization that prevents over-fitting. One situation where MTL may help is if the tasks share significant commonalities and are generally slightly under sampled[ \cite{hajiramezanali_bayesian_2018}].

In the context of Deep Learning, it's the most commonly approach for multi-task learning to use hard parameter sharing, generally applied by sharing the hidden layers between all tasks, while keeping task-specific output layers. MT-DNN model is a one of the typical hard parameter sharing application in NLU tasks\cite{liu_multi-task_2019}.

\subsection{Meta Learning}

Meta-learning, or learn-to-learn, has recently attracted much attention in the machine learning community \cite{lake_human-level_2015}. Basically the goal of meta-learning is to train a learner that is able to fast adapt to new task with limited training data. 

There are three common approaches to meta-learning: metric-based, model-based, and optimization-based. 
\paragraph{Metric-based}


Metric-based meta-learning is similar to nearest neighbors algorithm and kernel density estimation. The model predicts a probability y over a set of known labels by a weighted sum of labels of support set samples. The weight is generated by a kernel function $k_\theta$, which measures the similarity between two data samples.
$$P_\theta(y|x,S)=\sum_{(xi,yi) \in S} k_\theta(x,x_i)y_i$$ To train a successful metric-based meta-learning model requires researchers to specify a good kernel that could learn the distance function over objects well. However,  kernel functions are highly depends on specific problem, the inputs and the representation space of tasks.

Most of the frequently-used metric-based models learn embedding vectors of input data explicitly and use them to design proper kernel functions, see Convolutional Siamese Neural Network\cite{koch_siamese_nodate}, Relation Network \cite{sung_learning_2018}, etc.
\paragraph{Model-based}

Model-based meta-learning model requires a model designed for fast learning. 
By modifying the internal model architecture or adding an additional meta-learner model on top of original model, it could achieve the fast learning goal, i.e. to update its parameters rapidly within a few training steps.

The representative works in this category include Memory-Augmented Neural Network\cite{pmlr-v48-santoro16}, Meta Network\cite{Munkhdalai2017MetaN}, etc.
\paragraph{Optimization-based}

Optimization-based meta-learning algorithms aim to achieve the fast adaptation goal by adjusting the optimization algorithms. As we all know, deep learning models learn through back-propagation of gradients. Yet, since the gradient-based optimization does not work well on small number of training samples and won’t converge within a small number of optimization steps, a model is designed to modify the gradient based optimization algorithm.

The most popular optimization-based meta-learning algorithm is model-agnostic meta-learning(MAML), which is also what we mainly aim to investigate in this paper. \cite{finn_model-agnostic_2017} achieved state-of-the-art performance by directly optimizing the gradient towards a good parameter initialization for easy fine-tuning on low resource scenarios without introducing any additional architectures or parameters. 

\begin{figure}
    \centering
    \includegraphics[scale=0.4]{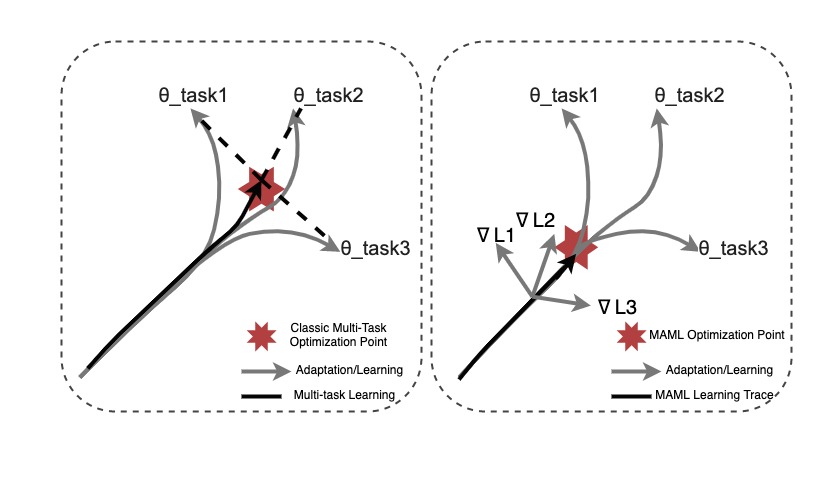}
    \caption{The illustrative comparisons between representations learned by (a) classical multi-task learning and (b) meta multi-task learning.}
    \label{fig: multi_vs_meta}
\end{figure}

Figure \ref{fig: multi_vs_meta} visually illustrates the differences between classical multi-task learning and meta multi-task learning. The classical multi-task learning tends to get to a point where the current gradients from different tasks are balanced, which may still result in over-fitting and tend to favor tasks with significantly larger amounts of data than others, while meta-learning aims to minimize the future loss of different task respectively.

\subsection{Financial NLU Applications}
\subsubsection{Financial Sentiment Analysis}
General sentiment analysis aims to extract people's opinions or tendency from language. Yet there is a key specialty in financial sentiment analysis that the purpose of financial sentiment analysis is usually targeted towards the market. Indeed, it usually aims to analyze the text data to facilitate understanding of how the markets will react with the information presented in the text. 

Most popular methods in solving sentiment analysis tasks include RNN, LSTM network models. Extending upon these models, \cite{8334488}  adds a text simplification layer and then applies it to LSTM network. Despite the success in general sentiment analysis, there is still a huge gap to utilize the neural networks to their fullest potential in solving tasks in finance domain due to the lack of high quality annotated datasets in the domain. 


 \cite{araci_finbert:_2019} has tackled this issue with FinBERT model. As we have discussed before, it essentially is to initialize the model with pre-trained values and fine-tuning the model with respect to the classification task. In FinBERT, the author used Reuters data to pre-training the BERT model and achieved promising results on Financial PhraseBank. 

We address this problem  from a different perspective. We utilize the meta learning model by training it with multiple NLU tasks to facilitate learning of a more robust and generalized representation. Then fitting the model to Financial PhraseBank dataset so that the model can quickly adapted on learning the sentimental relations in the text. We have compared our results against their reported accuracy.

\subsubsection{Stock Price Prediction}
Stock price prediction has long attracted researchers and investors. In financial natural language processing field, the two primary content resources for stock market prediction are public news and social media data (mainly from twitter). Classical research relies primarily on feature engineering but their results tends to be highly volatile. With the prevalence of deep learning\cite{le_distributed_nodate}, event driven approaches were studied and models with LSTM, RNN become dominant. More recently, \cite{hu_listening_2019} proposed a novel method to feed news sequence directly from text with hierarchical attention mechanisms for stock trend prediction. Further, language model such as BERT has also inspired researchers with development of new models.

However, stock price prediction is widely considered difficult due to three factors: high market stochasticity, chaotic market information and temporally-dependent prediction.  Stock prices are not influenced solely by news information or tweeter information. Other factors influencing the stock price are not directly observable or measurable. Thus the traditional prediction are mainly resulting in a random-walk pattern\cite{noauthor_random_1996}. 

In order to tackle the temporally-dependent prediction issue, researchers choose to frame the data to fit for a time series problem. In other word, they have to incorporate the temporal dependency between stock prices movements in to the model. For example,  when a company experienced some good news on day $d_1$, its stock price will be slowly affected and thus will have an upward trend pattern in the following days until $d_n$. Similarly, when a company suffered from some scandal, its stock price will needs time to absorb the affect of the scandal in the following n days.

Yet the time series model explained above did not address the chaotic market information issue. Different stock may correlated in different level with text data. Some stock may suffer from inefficient data issue. Transfer learning provides a viable way to alleviate this issue by using meta-learning. Previously, Zhaojiang has used MAML in tackling a similar issue: use text data from Chinese Weibo to predict sales for different company\cite{lin_learning_2019}.  This strategy used non-parametric model to leverage historical information of other brands, and used them as prior knowledge and thereby allows the model for fast adaptability.

In this paper, we are going to adopt the methodology of MAML to  test the effectiveness of MAML on stock price prediction task.

\section{Method}
In this section, we present the main methods used: pre-trained Language Model of BERT and Model-Agnostic Meta-Learning algorithm, and how we combine them.

\subsection{Pre-trained Language Model: BERT}
We use Bidirectional Encoder Representations from Transformers (BERT)\cite{devlin_bert:_2019} as our pre-trained model, which will be shared across all the tasks. 

BERT is first trained on unlabelled text, including Brown Corpus and English Wikipedia which has more than 2.5 billions of words. Fine-tuned on downstream nature language processing jobs, BERT has obtained state-of-art results on 11 different tasks, such as text classification, named entity recognition, sentiment analysis and question answering. Unlike ELMo which predicts the next word of an ordered sequence of tokens, BERT is trained on the entire sentence by randomly masking $15\%$ of the set of words. Therefore, instead of learning the context based on the previous or next word, it can learn the representation of words through all words in the sentence simultaneously. With transformer and bi-directional structure implemented, BERT helps with disambiguation of polysemous words and homonyms by focusing attention on a specific token. 

BERT has two versions: BERT-Base, with 12 encoder layers, hidden size of 768, 12 multi-head attention heads and 110M parameters in total, and BERT-Large, with 24 encoder layers, hidden size of 1024, 16 multi-head attention heads and 340M parameters. Considering about the computation resources, we only use BERT-Base in our experiments.

\subsection{Algorithm: Model-Agnostic Meta-Learning}
The basic idea of MAML\cite{finn_model-agnostic_2017} and its variants is to use a set of source tasks to find the initialization of parameters, and by using that parameters, it would require only a small number of training examples to learn a target task.

Given a set of tasks $\{\mathcal{T}_1,...,\mathcal{T}_k\}$ drawn from a distribution of $p(\mathcal{T})$, which consist of a training set $train(\mathcal{T})$ and a testing set $test(\mathcal{T})$, consider a model represented by a parameterized function $f_\theta$ with parameters $\theta$. 

When adapting to new tasks $\mathcal{T}_i$, we can update the model's parameter $\theta$ to $\theta_i'$ using one or more gradient update(We use one gradient update here to simplify the case, but usually real applications use multiple gradient updates):
$$ \theta_i'=\theta -\alpha \nabla_\theta \mathcal{L}_{\mathcal{T}_i}(f(\theta)) $$

This is the inner loop update, where $\mathcal{L}_{\mathcal{T}_i}$ is the loss function for $\mathcal{T}_i$.

To achieve a good generalization across various tasks, we aim to optimize the meta-objective, which is as follows:
$$ \min_{\theta} \sum_{\mathcal{T}_i \sim p(\mathcal{T} )} \mathcal{L}_{\mathcal{T}_i}(f(\theta')) = \sum_{\mathcal{T}_i \sim p(\mathcal{T} )} \mathcal{L}_{\mathcal{T}_i}(f(\theta -\alpha \nabla_\theta \mathcal{L}_{\mathcal{T}_i}(f(\theta)))) $$

We perform the meta-optimization over the model parameters $\theta$, with the objective computed using the updated model parameters $\theta$. As a result, a few gradient steps on a new task will produce maximum influence on that task.

So for the outer loop, model parameters $\theta$ are updated as follows: $$ \theta \leftarrow \theta - \beta \nabla_\theta \sum_{\mathcal{T}_i \sim p(\mathcal{T} )}\mathcal{L}_{\mathcal{T}_i}(f(\theta)) $$
where $\beta$ is the meta step size. The full algorithm is outlined in Algorithm \ref{alg:MAML}, adapted from \cite{finn_model-agnostic_2017}.

\begin{algorithm}[!h]
\SetKwInput{kwRequire}{Require}
\SetAlgoLined
\kwRequire{$p(\mathcal{T})$:distribution over tasks}
\kwRequire{$\alpha$,$\beta$: step size hyperparameters}
 randomly initialize $\theta$\;
 \While{not done}{
  Sample batch of tasks $\mathcal{T}_i \sim p(\mathcal{T} )$\;
  \ForAll{$\mathcal{T}_i$}{
    Evaluate $\nabla_\theta \mathcal{L}_{\mathcal{T}_i}(f(\theta)) $ with respect to K examples\;
    Compute adapted parameters with gradient descent: $\theta_i'=\theta -\alpha \nabla_\theta \mathcal{L}_{\mathcal{T}_i}(f(\theta))$\;
   }
   Update $\theta \leftarrow \theta - \beta \nabla_\theta \sum_{\mathcal{T}_i \sim p(\mathcal{T} )}\mathcal{L}_{\mathcal{T}_i}(f(\theta))$\;
 }
 \caption{Model-Agnostic Meta-Learning(MAML)}
 \label{alg:MAML}
\end{algorithm}

\subsection{Proposed Framework}

The architecture of the MAML model is similar to MT-DNN\cite{liu_multi-task_2019}. A word sequence (either a sentence or a pair of sentences packed together) is firstly input to BERT, which is the shared semantic representation trained by our meta multi-task objectives. On the top are the task-specific layers, where for each task, task-specific representations are generated by task-specific layers. And after that, there are some necessary operations for classification, relevance ranking, etc.

Generally, there are three steps in our method: the pre-training step as in BERT, the meta-learning step and fine-tuning step. In meta multi-task learning step, we use stochastic gradient descent (SGD) for inner loop update and Adamax optimizer for outer loop adaptation. In each epoch, a batch of tasks is selected, and the model is updated according to the sum of all multi-task objectives over the tasks.

\section{Experiments}

In this section, we discuss two specific instantiations of MAML for multi-task learning settings. One is multi-types of NLU tasks and another is multiple NLU tasks in single type, which differ in the loss function's form and in how data is generated by the tasks and presented to the model, but the same basic adaptation mechanism are applied in both cases.

\subsection{Multi-Types NLU Tasks}
In this part, we study the performance of MAML model with multiple types of tasks.

\subsubsection{Datasets}
We briefly describes the GLUE, SNLI, and SciTail datasets, as summarized in Table \ref{benchmarks}.
\paragraph{GLUE} 
The General Language Understanding Evaluation (GLUE) benchmark \cite{wang_glue:_2018} is a tool for evaluating and analyzing the performance of natural language understanding models across nine NLU tasks: Single-Sentence Tasks, Similarity and Paraphrase Tasks and Inference Tasks. 

Four high-resource datasets(MNLI, QQP, SST, QNLI) are used as training datasets, and four other low-resource datasets(CoLA, MRPC, STS-B, RTE) are used as testing datasets, according to \cite{dou_investigating_2019}. In our experiments we do not train or test models on the WNLI dataset because of previous work \cite{devlin_bert:_2019}.
\paragraph{SNLI} 
The Stanford Natural Language Inference dataset \cite{bowman_large_2015} is a naturalistic corpus of 570k sentence pairs labeled for entailment, contradiction, and independence. 

We use this dataset to examine the algorithm's fast adaptation ability in this study.

\paragraph{Sci-Tail}

This is a Textual Entailment Dataset from Science Question Answering \cite{khot_scitail:_nodate}. Hypotheses from science questions are created while the corresponding answer candidates and premises from relevant web sentences are retrieved from a large corpus. These linguistically challenging sentences, combined with the high lexical similarity of premise and hypothesis for both entailed and non-entailed pairs, makes the new entailment task particularly difficult. 

We use this dataset examine the algorithm's fast adaptation ability in this study.

\paragraph{Financial PhraseBank(FPB)}
The sentiment analysis dataset \cite{malo_good_2013} consists of 4845 english sentences selected randomly from financial news found on LexisNexis database, which is annotated by 16 people with finance and business background. The annotators were asked to give labels according to how they think the information in the sentence might
affect the mentioned company stock price. 

This dataset is our first step to generalize our model to financial domain in this study.

\begin{table}[!h]
  \centering
  \begin{tabular}{lllllll}
    \toprule
    \cmidrule(r){1-7}
    Corpus     &Task    & Train     & Dev    & Test     & Label  &Metrics \\
    \midrule
    CoLA     & Acceptability     & 8.5k    & 1k &1k     & 2  &Matthews Corr  \\
    SST-2     & Sentiment & 67k &872    &1.8k   & 2 &Accuracy     \\
    MNLI     & NLI     & 393k    & 20k &20k     & 3  &Accuracy(match/mismatch) \\
    RTE     & NLI  & 2.5k &276    &3k   & 2 &Accuracy    \\
    WNLI    &NLI &634 &71 &146 &2 &Accuracy \\
    QQP     & Paraphrase & 364k &40k    &391k   & 2 &Accuracy/F1 \\
    MRPC     & Paraphrase & 3.7k &408    &1.7k   & 2 &Accuracy/F1     \\
    STS-B     & Similarity     & 7k    & 1.5k &1.4k     & 1  &Pearson/Spearman Corr\\
    \midrule
    QNLI     & QA/NLI     & 108k    & 5.7k &5.7k     & 2  &Accuracy \\
    SNLI     & NLI      & 549k    & 9.8k &9.8k     & 2  &Accuracy\\
    SciTail     & NLI     & 23.5k    & 1.3k &2.1k     & 2  &Accuracy \\
    \midrule
    FPB &Sentiment & 2.9k& 1.0k &1.0k &3 &Accuracy\\
    \bottomrule
  \end{tabular}
  \bigskip
   \caption{Benchmarks: GLUE, SNLI, SciTail, FPB}
  \label{benchmarks}
\end{table}

\subsubsection{Implementation Details}

Our implementation of MAML is based on PyTorch implementation of MT-DNN \cite{liu_multi-task_2019}. We use Adamax with a learning rate of 5e-5 as our outer optimizer, batch size of 32 and the maximum number of epochs of 5. We also set the update step to 3 and $\alpha$, the inner learning rate of SGD to 5e-5. The dropout rate of all task specific layers is 0.1, except 0.3 for MNLI and 0.05 for CoLA. A linear learning rate decay schedule with warm-up over 0.1 is used. The gradient norm is clipped within 1 to avoid exploding gradient problem. The pre-trained BERT-Base is used to initialize the model. Tasks are sampled according to the size of their datasets. 

An Amazon p3.8.xlarge EC2 instance with 4 GPUs, and 90 GiB of host memory is used to train the models. 

\begin{figure}[!h]
    \centering
    \includegraphics[scale=0.35]{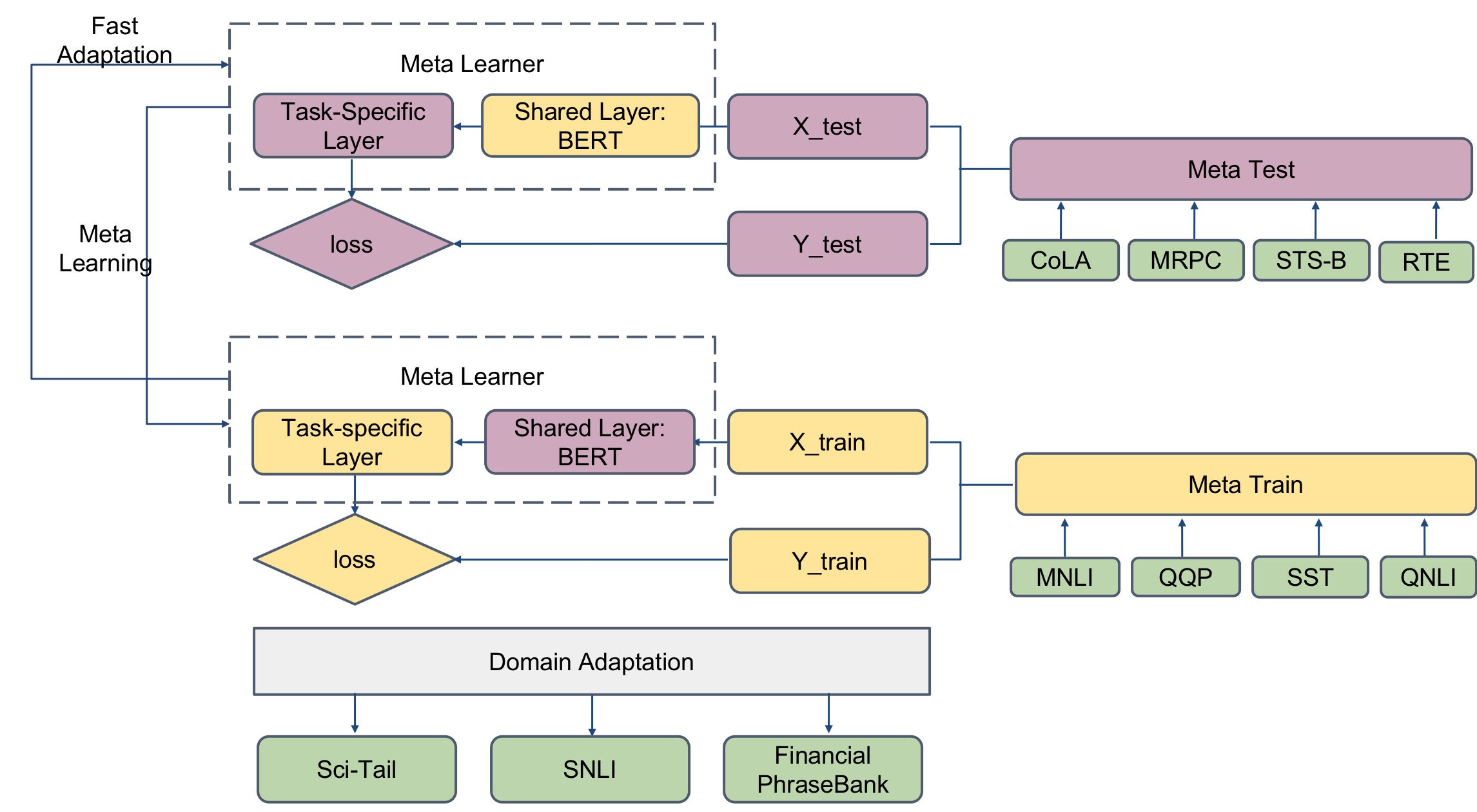}
    \caption{Multi-types NLU Tasks Learning}
    \label{fig: multitype}
\end{figure}

Experiment details are presented in Figure \ref{fig: multitype}. 
We compare our MAML model against various state-of-the-art baselines. 

For GLUE, SciTail and SNLI datasets, We use the public code of BERT-Base \cite{devlin_bert:_2019} model and MT-DNN model\cite{liu_multi-task_2019} to obtain their results. For Financial PhraseBank dataset, we target the results in FinBert model\cite{araci_finbert:_2019}.

\subsubsection{Results}
The experiment results on GLUE, SNLI, SciTail and Financial PhraseBank datasets are the following.
\paragraph{GLUE Main Results} We first train the MAML model using four of the GLUE datasets and their fine-tuned results are presented in Table \ref{Glue Train Results}. Then we test the model with four GLUE datasets. The results for the testing datasets are presented in Table \ref{Glue Test Results}. 

Basically, the MAML model achieves better or equal performance in almost all tasks, which indicates the effectiveness and reliability of our model.
\begin{table}[!h]
  \centering
  \begin{tabular}{lllll}
    \toprule
    \cmidrule(r){1-5}
    Model     &MNLI    & QQP     & SST   & QNLI \\
    \midrule
    BERT     & 84.6/83.4     & 71.2    & 93.5 & 90.5  \\
    MT-DNN     & 84.3/84.5     & 86.9    & 92.9 & 90.8  \\
    MAML    & 84.0/84.4 & 87.1 &92.7    &90.5  \\
    \bottomrule
  \end{tabular}
  \bigskip
  \caption{\label{Glue Train Results}Training Results on GLUE Datasets. MAML and MT-DNN uses BERT-Base to initialize their shared layers. We fine-tuned three models for each of the four GLUE task using task-specific data.}
\end{table}
\begin{table}[!ht]
  \centering
  \begin{tabular}{lllll}
    \toprule
    \cmidrule(r){1-5}
    Model     &CoLA    & MRPC     & STS-B   & RTE \\
    \midrule
    BERT     & 52.1     & 84.8/88.9    & 66.4 & 87.1/85.8  \\
    MT-DNN     & 55.9 & 87.2/90.5 & 74.4    &89.6/89.6  \\
    MAML    & 56.9 & 87.3/90.7 &78.3    &89.3/89.3  \\
    \bottomrule
  \end{tabular}
  \bigskip
  \caption{\label{Glue Test Results}Testing Results on GLUE Datasets. MAML and MT-DNN uses BERT-Base to initialize their shared layers. We fine-tuned three models for each of the four GLUE task using task-specific data.}
\end{table}

\paragraph{Fast Adaptation on SNLI and SciTail} We transfer our model to two new tasks. We randomly sample 0.1\%, 1\%, 10\% and 100\% of their training data and thus obtain four sets of training data for Sci-Tail including 23, 235, 2.3k, 23.5k training samples, and four sets for SNLI including 549, 5.5k, 54.9k and 549.3k training samples respectively.

We observe that MAML model outperforms the BERT and MT-DNN baselines with fewer training examples used, with more details provided in Table \ref{Domain Adaptation}. For example, with only 0.1\% of the Sci-Tail training data, MAML achieves an accuracy of 77.531\% while BERT's is 51.2\% and MT-DNN's is 66.411\%. 

Similar results are also verified in SNLI dataset.

\begin{table}[!h]
  \centering
  \begin{tabular}{lllll}
    \toprule
    \multicolumn{5}{c}{SNLI(Dev Accuracy $\%$)}                   \\
    \midrule
    $\#$Training Data     & 0.1$\%$    & 1$\%$     & 10$\%$    & 100$\%$  \\
    BERT     & 52.5     & 78.1    & 86.7 & 91.0  \\
    MT-DNN     & 81.6 & 84.7 & 88.0    &91.08  \\
    MAML    & 82.0 & 84.9 &88.2    &91.4  \\
    \midrule
    \multicolumn{5}{c}{Sci-Tail(Dev Accuracy $\%$)}                   \\
    \midrule
    $\#$Training Data     & 0.1$\%$    & 1$\%$     & 10$\%$    & 100$\%$  \\
    BERT     & 51.2     & 82.2    & 90.5 &94.3   \\
    MT-DNN     & 66.411  & 90.874 &92.638    &94.862 \\
    MAML     & 77.531 & 89.801 &91.648    &95.015  \\
    \bottomrule
  \end{tabular}
  \bigskip
  \caption{\label{Domain Adaptation}Domain Adaptation Results on SNLI and Sci-Tail using the shared embeddings generated by MAML, MT-DNN and BERT, respectively.}
\end{table}

\begin{figure}[!h]
  \centering
  \includegraphics[scale=0.6]{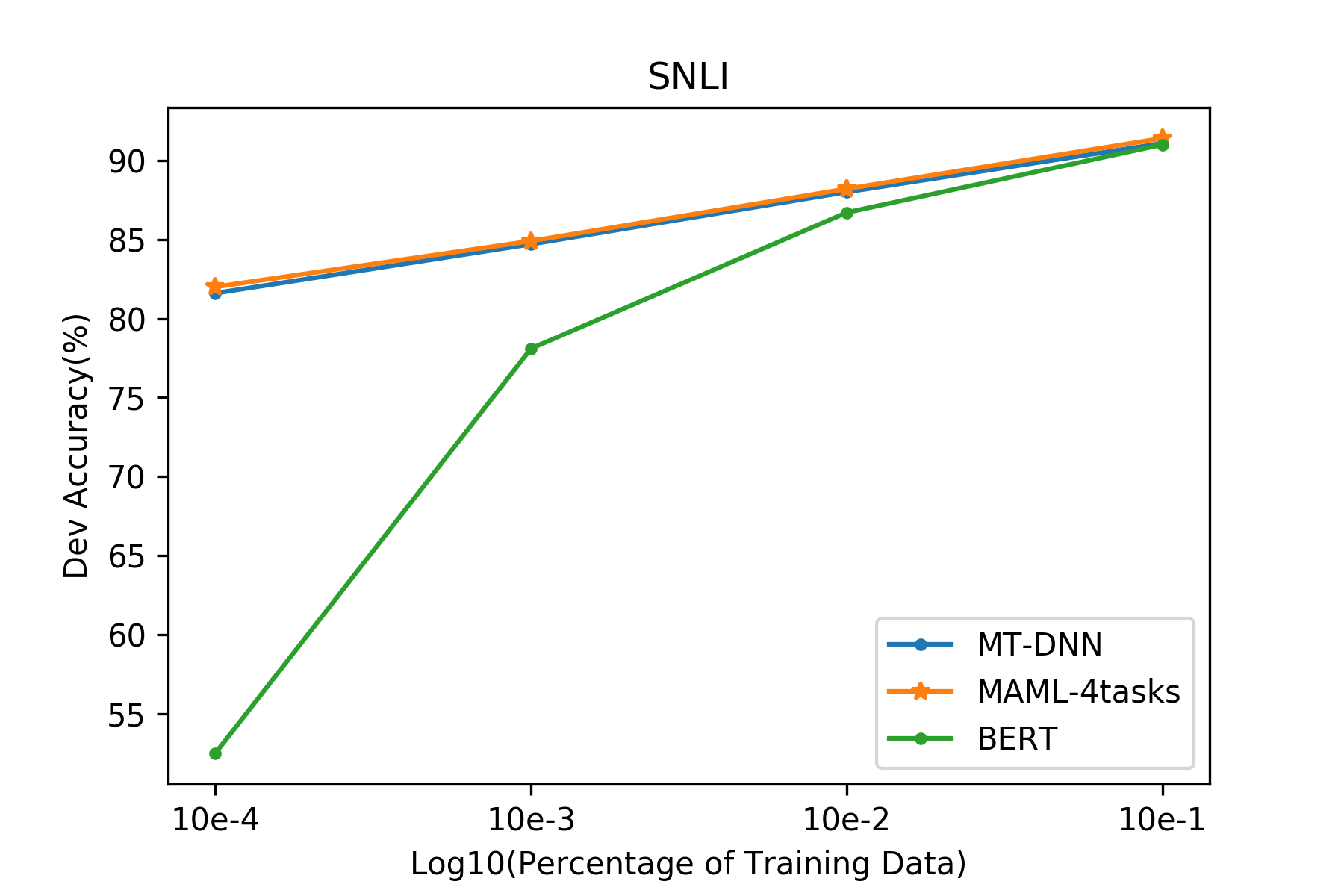}
  \caption{Results on SNLI Dataset. The X-axis indicates the amount of domain-specific labeled samples used for adaptation.}
\end{figure}
\begin{figure}[!h]
  \centering
  \includegraphics[scale=0.6]{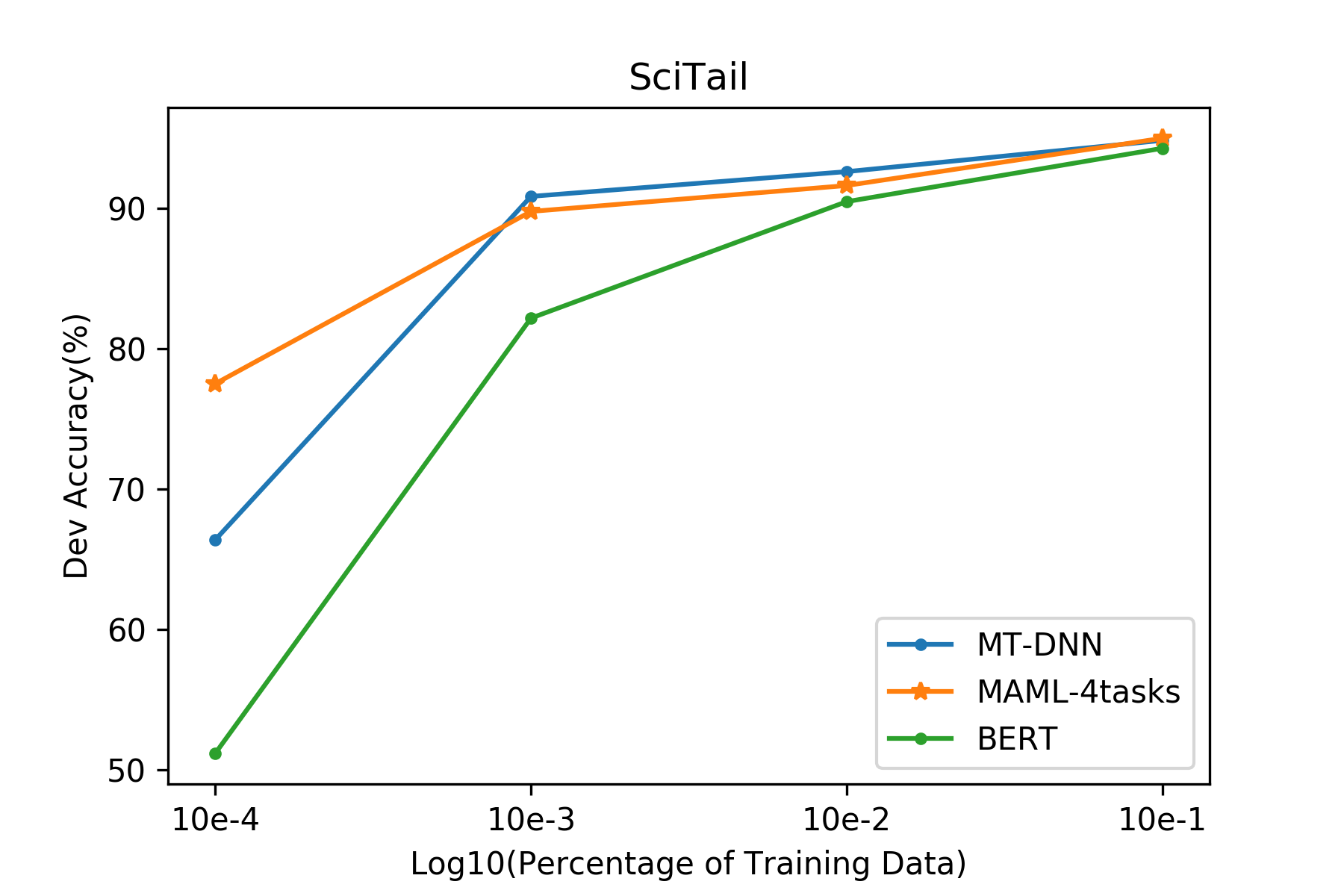}
  \caption{Results on Scitail Dataset. The X-axis indicates the amount of domain-specific labeled samples used for adaptation.}
\end{figure}

\paragraph{Domain Adaptation on Financial Dataset} Based on above experiments, we further extend our model to Financial PhraseBank dataset. From Table \ref{FPB}, we could see that the MAML model achieves an accuracy as good as FinBert, without any financial specific further pre-training.  Moreover, with only 1\% or 10\% training data, it reaches a fairly good performance.

\begin{table}[!h]
  \centering
  \begin{tabular}{llll}
    \toprule
    \multicolumn{4}{c}{FPB(Dev Accuracy $\%$)}                   \\
    \midrule
    $\#$Training Data   & 1$\%$     & 10$\%$    & 100$\%$  \\
    Fin-Bert & - & -&86.00 \\
    MAML    &61.26 &  77.38& 86.47 \\
    \bottomrule
  \end{tabular}
  \bigskip
  \caption{\label{FPB}Domain Adaptation Results on Financial PhraseBank.}
\end{table}

\subsection{Single-Type NLU Tasks - Stock Price Prediction}
The experiments above show the effectiveness of MAML in handling multiple tasks together. In this part, we aim to apply MAML to solve single-type financial NLU task, stock price prediction.
\subsubsection{Dataset}
We obtain the dataset from \cite{xu_stock_2018}. There are two main components in our dataset, a Twitter dataset and a historical price dataset. The historical prices for the 88 selected stocks to build the historical price dataset from Yahoo Finance. the text data  includes two-year price movements from 01/01/2014 to 01/01/2016 of 88 stocks separated into 9 industries: Basic Materials, Consumer Goods, Healthcare, Services, Utilities, Conglomerates, Financial, Industrial Goods and Technology. The table blow shows that there is an imbalance issue lies within stocks and industries which we have to deal with in training/evaluation phase.

\begin{table}[!h]
  \centering
  \begin{tabular}{llll}
    \toprule
    \multicolumn{4}{c}{Number of Twitter Text Per Industry}                   \\
    \midrule
    $\#$Industry   & \#Num Twitter Text  \\
    Material & 4405\\
    Consumer Goods &22491\\
    Healthcare & 7984 \\
    Services     & 19025 \\
    Utilities     & 6095 \\
    Cong     & 268 \\
    Finance     &9291 \\
    Industrial Goods     &  5764\\
    Tech     & 31015 \\
    \bottomrule
  \end{tabular}
  \bigskip
  \caption{\label{Financial Adaptation 1}Number of Twitter Text Per Industry During 01/01/2014 - 01/01/2016.}
\end{table}

\subsubsection{Implementation-Single Stock Price Prediction Task}

  Inspired by \cite{xu_stock_2018}, we assume that predicting the stock movement between trading day d and d+1 can be benefit from historical prices of previous days and previous price movements on its former trading days. Under this premise, we adopt the data processing techniques from \cite{xu_stock_2018}. First, we find all T eligible trading days referred in a sample stock and group them by $t \in [1,T]$. Thus each sample should contain twitter text and stock price data with in the range of t days. Let us use $S = [s_1, s_2,... , s_t] , P = [p_0,p_2,...p_t] $ to represent the twitter text collected in each sample which collected by aligning to each trading day. Then we transform the text data and stock price in to the features we desired. We calculate stock price movement $Y=[p_1-p0,p_2-p_1,...,p_t-p_{t-1}]$. Note here we further transform the Y into three classes: moving up, moving down, no movement.

The architecture of the model is shown below. We first use BERT to process the twitter data and concat it with the previous days stock price. Then we feed it into a RNN model with T layers which represent T days in the lag. Finally, we integrate the final result with a softmax function in order to output the confidence distribution over up and down. 	 
\begin{figure}[!h]
  \centering
  \includegraphics[scale=0.55]{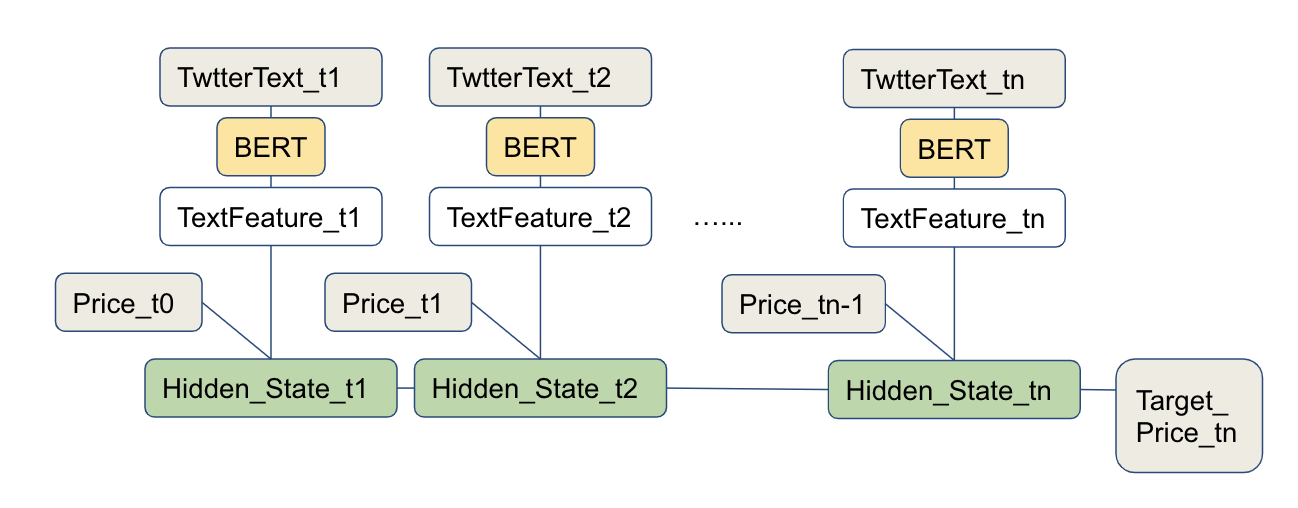}
  \caption{The architecture of Single Stock Prediction Task. We use the main target tn for prediction and the lag size of n for illustration}
\end{figure}

\subsubsection{Implementation- MAML-BERT Model}
 We then transform the model to multitask structure by adding multiple tasks together and applying MAML method to it. The scenario is, suppose we are given a new stock with limited twitter text data, with MAML model pre-trained on multiple stock-text data, the model should be able to capture the intrinsic parameters for this new stock quicker and thus reach good accuracy faster. 
 
\begin{figure}[!h]
  \centering
  \includegraphics[scale=0.40]{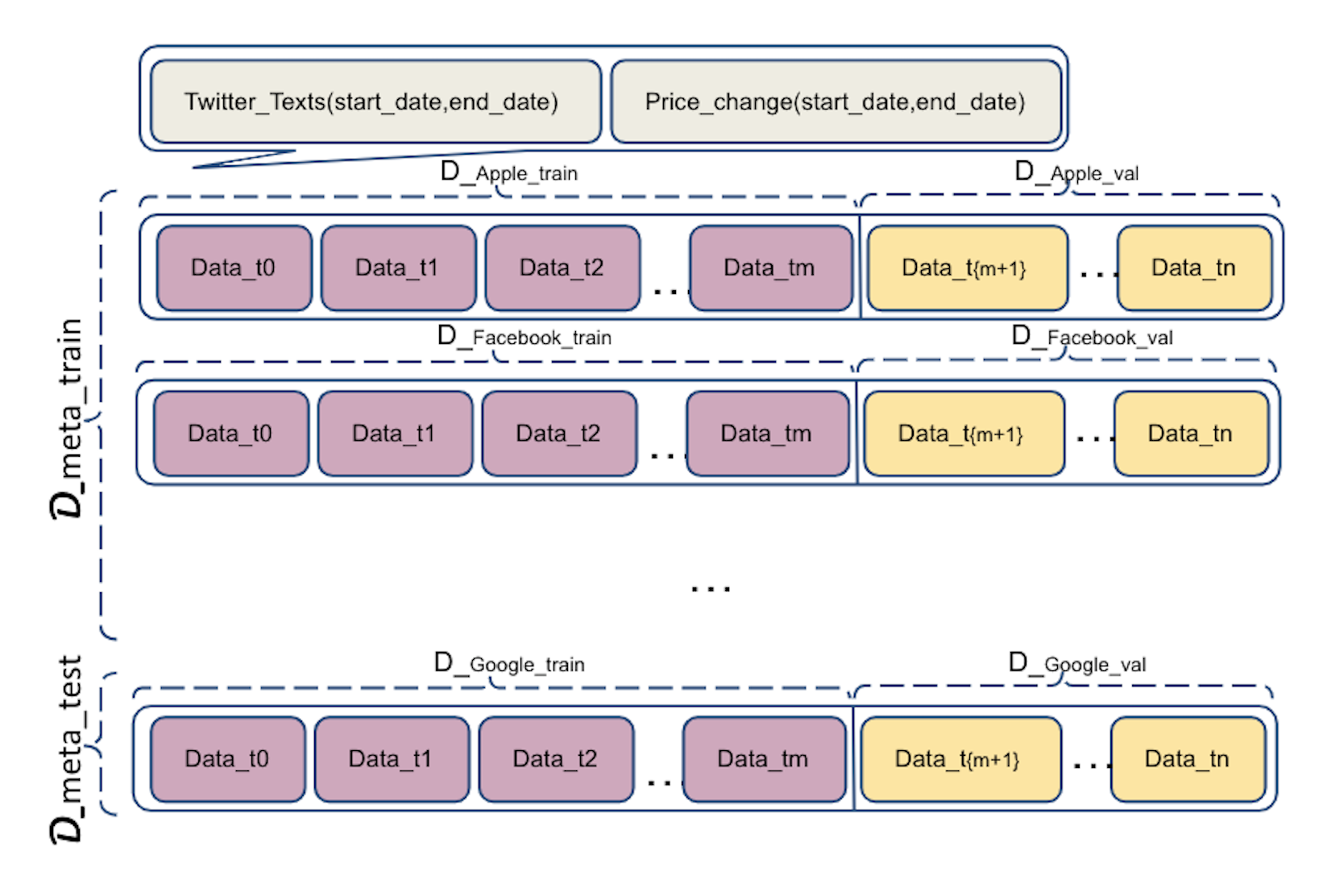}
  \caption{\label{structure}MAML-Model task structure}
\end{figure}

To test our hypothesis scenario, we design experiments by selecting 8 stocks to train the meta learner for 10 epoch and test the model with a new stock dataset. We conduct 4 experiments and finally evaluate its accuracy against a direct prediction model listed in 4.2.2. The setting structure is shown in Figure \ref{structure}.

\subsection{Results}
As described in previous sections, stock prediction is a challenging task and
a minor improvement could lead to large potential profits. An the accuracy of 56\% is generally reported as a satisfying result for binary stock movement prediction\cite{nguyen_topic_2015}. We evaluate the model in the following four settings and the results are illustrated in the graph below. The selection of the stocks is according the the amount of twitter text data we obtain.

\begin{table}[!h]
  \centering
  \begin{tabular}{lllll}
    \toprule
    \cmidrule(r){1-5}
    Model Industry     &Train Stocks    & Eval Stock     & (Train Acc $\%$)   & (Eval Acc$\%$) \\
    \midrule
    Consumer     &PG,BUD,KO,PM,TM,PEP,..     &AAPL    & 59.21 & 57.14  \\
    Services    & AMZN,BABA,WMT,CMSCA,..    & MCD    & 58.82 & 56.91  \\
    Tech    & GOOG, MSFT,FB,T,CHL,ORCL,.. & CSCO &58.42    &56.15  \\
    Mixed    &CELG,PCLN,JPM,GE,FB.. & AAPL &60.27    &57.94  \\
    \bottomrule
  \end{tabular}
  \bigskip
  \caption{\label{Train Results}Training and Evaluation Results on Different Groups of Stock Data.}
\end{table}

As shown in the table and graph below, we have reached promising evaluation accuracy on all different models. The highest result is generated by Mixed Model. It is trained on 8 stocks with the maximum number of twitter text data from all industry and evaluated on AAPL stock which is not included in the industry for all training stock.  The final dev accuracy we have reached is of 57.94\% for MAML-mixed model.

We have also evaluate the adaptation rate of our model(MAML-Mixed) against a baseline model that was not pre-trained on other stocks. The graph below shows that the MAML model converged in a faster rate compared to the baseline model.
\begin{table}[!h]
  \centering
  \begin{tabular}{llll}
    \toprule
    \multicolumn{4}{c}{Stock Price Prediction (Dev Accuracy $\%$)}                   \\
    \midrule
    RAND & 50.89\\
    ARIMA &51.39\\
    BERT-Baseline & 54.07 \\
    BERT-MAML-Mixed     & 57.94 \\
    Stocknet (\cite{xu_stock_2018}) & 57.64\\
    \bottomrule
  \end{tabular}
  \bigskip
  \caption{\label{Financial Adaptation 2}Domain Adaptation Results on Stock Price Prediction Tasks.}
\end{table}

\begin{figure}[!h]
  \centering
  \includegraphics[scale=0.65]{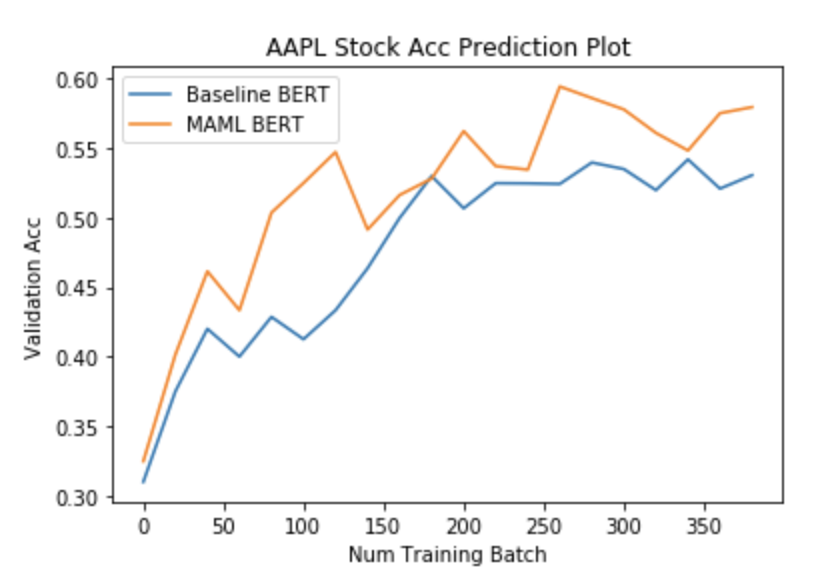}
  \caption{Results on Stock Price Prediction Task from BERT Model and BERT-MAML model}
\end{figure}

\section{Conclusions}
In this paper, we investigate model-agnostic meta-learning algorithm for general NLU tasks, and also evaluate its performance on two financial applications - financial sentiment analysis and stock price prediction. Experiments show our MAML model is able to learn general representations, which can be adapted to new tasks with limited samples effectively, and is also robust to the task specific scales without over-fitting. Our study suggests promising applications of meta-learning algorithms in low-resource financial natural language understanding tasks.

\section*{Acknowledgments}
The authors are grateful to the project advisers from Center for Data Science and Dr. Zulkuf Genc, Dmitri Jarnikov and Dogu Araci from Prosus AI team for their constructive comments and advisement. The computational resources are provided by Prosus AI team.

\bibliographystyle{unsrt}  
\bibliography{references}

\end{document}